\documentclass[runningheads]{llncs}

\usepackage[american]{babel}
\usepackage[T1]{fontenc}
\usepackage[utf8]{inputenc}% For umlauts (OT1) and the ß (OT1, T1).
\usepackage[hang,flushmargin]{footmisc}
\usepackage{xurl}
\usepackage[hidelinks]{hyperref}

\usepackage{type1cm}
\renewcommand{\tt}{\fontencoding{OT1}\fontfamily{cmtt}\selectfont\small}
\newcommand{\Ni}{(1)~}
\newcommand{\Nii}{(2)~}
\newcommand{\Niii}{(3)~}
\newcommand{\Niv}{(4)~}
\newcommand{\Nv}{(5)~}

\usepackage{booktabs}
\usepackage{multirow}
\usepackage{graphicx}
\DeclareGraphicsExtensions{.pdf,.ai,.jpg,.png}
\setkeys{Gin}{pagebox=artbox}
\usepackage[numbers,sectionbib]{natbib}%[round]
\newif\ifbscomment
\bscommentfalse
\bscommenttrue% Show comments.

\RequirePackage{xcolor}
\RequirePackage{soul}
\setstcolor{blue}

\newsavebox\bscombox
\newcommand{\bscom}[3][]{%
  \sbox{\bscombox}{\fontsize{8}{9}\selectfont#1#2#3}
  \noindent
  \st{#2}{\selectfont
    \color{blue}#3\ifx\\#1\\\else{\color{violet}[#1]}\fi
    }
  }

\begin{document}

\title{\texorpdfstring{%
Overview of PAN 2026: \newline
Voight-Kampff Generative AI Detection, \newline
Text Watermarking, \newline
Multi-Author Writing Style Analysis, \newline
Generative Plagiarism Detection, and \newline
Reasoning Trajectory Detection}{%
Overview of PAN 2026: Voight-Kampff Generative AI Detection, Text Watermarking, Multi-Author Writing Style Analysis, Generative Plagiarism Detection, and Reasoning Trajectory Detection%
}}

% \subtitle{Extended Abstract}
\titlerunning{Overview of PAN 2026: Extended Abstract}
% \thanks{Authors are listed in alphabetical order.}

\author{
Janek Bevendorff,$^{1}$
Maik Fr{\"o}be,$^{2}$
Andr{\'e} Greiner-Petter,$^{3}$
Andreas Jakoby,$^{1}$
Maximilian Mayerl,$^{4}$
Preslav Nakov,$^{5}$
Henry Plutz,$^{6}$
Martin Potthast,$^{6,7,8}$
Benno Stein,$^{1}$
{Minh Ngoc} Ta,$^{5}$
Yuxia Wang,$^{9}$
Eva Zangerle$^{10}$
}
\authorrunning{Bevendorff et al.}

\institute{
$^{1}$Bauhaus-Universit{\"a}t Weimar, Weimar, Germany, \\
$^{2}$Friedrich Schiller University Jena, Jena, Germany, \\
$^{3}$Georg-August-Universit{\"a}t, G{\"o}ttingen, Germany, \\
$^{4}$University of Applied Sciences BFI, Vienna, Austria, \\
$^{5}$Mohamed bin Zayed University of Artificial Intelligence, Abu Dhabi, UAE, \\
$^{6}$University of Kassel, Kassel, Germany, \\
$^{7}$hessian.ai, Darmstadt, Germany, \\
$^{8}$ScaDS.AI, Leipzig, Germany, \\
$^{9}$INSAIT, Sofia University ``St. Kliment Ohridski'', Sofia, Bulgaria, \\
$^{10}$University of Innsbruck, Innsbruck, Austria \\[1.5ex]
\href{mailto:pan@webis.de}{\ttfamily pan@webis.de}\qquad \href{https://pan.webis.de}{\ttfamily pan.webis.de}
}

\maketitle

\begin{abstract}
The goal of the PAN workshop is to advance computational stylometry and text forensics via objective and reproducible evaluation. In 2026, we run the following five tasks:
\Ni~\emph{Voight-Kampff Generative AI Detection}, particularly in mixed and obfuscated authorship scenarios,
\Nii~\emph{Text Watermarking}, a new task that aims to find new and benchmark the robustness of existing text watermarking schemes,
\Niii~\emph{Multi-author Writing Style Analysis}, a continued task that aims to find positions of authorship change,
\Niv~\emph{Generative Plagiarism Detection}, a continued task that targets source retrieval and text alignment between generated text and source documents,
and \Nv~\emph{Reasoning Trajectory Detection}, a new task that deals with source detection and safety detection of LLM-generated or human-written reasoning trajectories.
As in previous years, PAN invites software submissions as easy-to-reproduce Docker containers for most of the tasks. Since PAN~2012, more than 1{,}100 submissions have been made this way via the TIRA experimentation platform~\cite{froebe:2023}.
\end{abstract}

\newpage

\section{Introduction}

The PAN lab is a workshop series and a networking initiative for stylometry and digital text forensics. We host shared tasks on authorship analysis, generative AI detection and analysis, and computational ethics in general. Since the workshop's inception in 2007, we have organized 82~shared tasks%
\footnote{Find PAN's past shared tasks at {\url{https://pan.webis.de/shared-tasks.html}}} %
(including this year's) and have assembled many original evaluation datasets\footnote{Find PAN's datasets at {\url{https://pan.webis.de/data.html}}} for this purpose. 

In 2025, our four tasks concluded with 112 software submissions and 70 notebook papers. %
Together with the ELOQUENT Lab, we continued the \textit{Voight-Kampff Generative AI Detection}%
\footnote{The name is inspired by the 1982 science fiction film Blade Runner. The Voight-Kampff machine is used to determine whether an individual is a human or a replicant in a polygraph-like test.}
task to develop methods for the reliable and robust detection of generative AI in the presence of adversarial text modifications. As in the 2024 installment, the task proved very successful and attracted a large number of submissions. This task will return in 2026. The 2025 edition was extended with another subtask to determine the degree of human-AI collaboration in a text, which also received many submissions, but will not return this year. %
The \textit{Multilingual Text Detoxification} task was continued as well with more languages in the dataset. Although quite successful, this task will not return in 2026, as we are taking our time analyzing the lessons learned from the previous editions. %
In its place, we introduce a new \textit{Text Watermarking} task aiming at developing robust watermarking techniques for existing text. %
As in previous years, the \textit{Multi-Author Writing Style Analysis} task attracted a stable number of participants, which we will build on again in~2026. %
The new \textit{Generated Plagiarism Detection} focused on the detection of near-verbatim text reuse by LLMs. Although it did not receive many submissions in 2025, we will continue this task in 2026.
Finally, we introduce \textit{Reasoning Trajectory Detection} as another new task in 2026, which has the goal of attributing reasoning trajectories to LLM or human authors and to classify their safety.

We briefly outline the upcoming tasks in the sections that follow.

\section{Task 1: Voight-Kampff Generative AI Detection}

With generative AI being ubiquitous, we now have the ability to produce high-quality discursive texts on virtually any topic, approaching human-like standards of writing. On the one hand, this is a notable achievement, but on the other hand, it is also a cause for concern. Recognizing the ``fingerprint'' of AI text generation is the foundation for a healthy information ecosystem in the future, but also to prevent model collapse when LLMs are trained increasingly on their own output. However, LLM detection remains a challenge, especially across text domains, but also in the face of potential LLM style obfuscations.

The Voight-Kampff task was conceived in 2024 as a successor to a long and well-received series of authorship verification tasks. Both the 2024 and the 2025 editions attracted many participants to submit their systems on the TIRA platform~\cite{froebe:2023}. The 2025 edition focused on robustness analysis of detectors against unknown text modifications (obfuscations) that try to mask the telltale LLM fingerprints. It was organized together with the ELOQUENT lab in a builder--breaker style, in which ELOQUENT participants contributed obfuscated datasets trying to break the PAN participants' systems. For 2026, we continue this line of research and explore new ways to make and break successful LLM detectors.

\section{Task 2: Text Watermarking}

Generative AI Detection has its limits~\cite{bevendorff:2025b}, which is why many AI companies now embed invisible watermarks into the output of their LLMs. The field of text watermarking is much older than LLMs~\cite{kaur:2015}, but as a result of the widespread adoption of generative AI, it has recently received renewed attention and novel methods have been developed. With \textit{Text Watermarking}, we propose a new task that is closely related to the Voight-Kampff task, but instead of detecting the style of an LLM, participants embed a watermark into an existing text and---in a second step---verify its existence. Between these two steps, the text goes through a number of unknown automated text obfuscation processes with the goal to destroy the embedded watermarks. Since the task uses existing text and does not rely on LLMs specifically, the submitted watermarking systems can be used in a much broader context to authenticate any type of text, not only machine-generated text. With the TIRA platform~\cite{froebe:2023} we are in a unique position to organize such a two-step task: the submitted software systems are executed in a stateless sandbox on our infrastructure, which prevents knowledge leaks between the steps. The systems are evaluated with regard to the inconspicuousness of the watermark and its robustness against (unknown) text modifications.

\section{Task 3: Multi-Author Writing Style Analysis}
The multi-author writing style analysis task aims to analyze the writing style of individuals in order to identify points where the style---and thereby the authorship---changes. This task is fundamental for several downstream applications, including intrinsic plagiarism detection and authorship verification.

The task was first introduced at PAN in 2016 and has since then witnessed steady progress and improvements. Initially, the focus was on identifying authors and grouping text segments by author~\cite{stein:2016i}. In the following years, the task was reformulated as a binary classification problem: detecting whether a text was authored by a single author or by multiple authors~\cite{kestemont2018overview,tschuggnall2017overview,zangerle2019overview}. In 2020, the task shifted to detecting whether there is a style change across all pairs of consecutive paragraphs~\cite{zangerle:2020} and to assigning authors to paragraphs~\cite{zangerle:2021, zangerle:2022}. In 2023 and 2024, the task again focused on the paragraph level, this time explicitly controlling for simultaneous changes in both authorship and topic~\cite{zangerle:2023,zangerle:2024}.

To ensure consistency and comparability across years, and thereby enable tracking of improvements, the PAN'26 edition will adhere to the established intrinsic multi-author writing style analysis task definition: ``\emph{For a given text, identify all positions where the writing style changes.}'' To solve the task, participants are asked to develop profiling methods capable of detecting whether a change in style---and thus authorship---occurs at specific positions in the text. However, in 2026, we will introduce a new dataset: We will make use of an extensive collection of fanfiction texts, i.e., user-written stories that reuse characters and settings from existing works. This allows us to extract longer, topically coherent excerpts while still maintaining control over authorship boundaries. Based on these input texts, we will apply our established mixing method to create the dataset underlying the task. To do so, we compute both a semantic and a stylistic representation of each text chunk, thereby enabling the construction of three datasets that differ in topical and stylistic similarity.

\section{Task 4: Generative Plagiarism Detection}

The widespread adoption of large language models (LLMs) has introduced complex challenges to modern academic workflows~\cite{CrothersJV23}, particularly with regard to academic integrity.
While several major scientific conferences have updated their content policies in recent years to allow (mostly partial) LLM-generated content~\cite{AAAIAiPolicy,AclAiPolicy,IcmlAiPolicy}, or to incorporate LLM-based assessments in their peer review processes,%
\footnote{\resizebox{.989\textwidth}{!}{\url{https://aaai.org/aaai-launches-ai-powered-peer-review-assessment-system/}}}
the boundary for what constitutes acceptable generated content is becoming increasingly blurred.%
\footnote{\url{https://www.intology.ai/blog/zochi-acl}}
As a result, identifying disingenuous research publications has become more difficult, as the methods for detecting LLMs and textual overlap are less relevant for modern plagiarism detection techniques.

The first revival of the plagiarism detection task at PAN in 2025 was successful, but lacked realism in terms of how LLMs might be used to create real-world plagiarism.
In the previous iteration, we introduced plagiarism paragraph-by-paragraph into an otherwise genuine publication.
While this approach resembled an unrealistic scenario, it was also relatively easy to detect the replaced paragraphs using simple embedding similarity calculations.
Moreover, as textual overlap decreases, the alignment sub-task of plagiarism detection becomes less relevant (i.e., identifying the precise boundaries of plagiarized paragraphs is less meaningful with substantial paraphrasing capabilities).

Therefore, we will develop a new version of the dataset, focusing on generating plagiarism from the ground up based on one or more source documents.
This will allow us to reinstate the crucial retrieval aspect of plagiarism detection and incorporate important cases of merging (summarization of multiple sources) and expanding plagiarism (splitting single sources into multiple plagiarized paragraphs).

To further enhance the dataset's relevance, we will also expand the domains covered by incorporating publications from PubMed\footnote{\url{https://huggingface.co/datasets/ncbi/pubmed}} (which includes the medical and chemical fields) and JSTOR\footnote{\url{https://support.jstor.org/hc/en-us/articles/32487330092695-JSTOR-Text-Analysis-Support-Working-with-JSTOR-Full-Text-Datasets}} (which spans the humanities and social sciences).
At the same time, the planned structure of the subtasks will largely mirror that of previous iterations.

\textit{Subtask 1 -- Source Retrieval:} 
This is the classic retrieval task for plagiarism detection systems.
Given a suspicious document and a corpus of potential source documents, the participant's system must identify all the source documents from the corpus or determine that no source documents are present if the suspicious document is authentic (i.e., the corpus contains no relevant sources for the document).

\textit{Subtask 2 -- Text Alignment:} 
This task focuses on aligning plagiarized paragraphs in the suspicious document with their corresponding paragraphs in the source documents. One plagiarized paragraph may correspond to multiple source paragraphs (merging plagiarism). Conversely, several plagiarized paragraphs might correspond to one source paragraph (expanding plagiarism).

\section{Task 5: Reasoning Trajectory Detection}

The year 2025 has witnessed substantial advancements in the reasoning capabilities of LLMs, improving both overall performance and safety through explicit reasoning trajectories before final answer~\cite{deepseekr1}. However, emerging evidence indicates that spurious, non-logical, or unsafe intermediate steps can still lead to incorrect or harmful final answers~\cite{mou2025saro}. In some cases, reasoning models may sometimes arrive at safe conclusions via deceptive or misaligned reasoning paths~\cite{frontier2024scheming}. To mitigate unsafe LLM reasoning, we propose a new task, \textit{Reasoning Trajectory Detection}, to deepen our understanding of AI-generated reasoning and support future improvements in reasoning and safety with two related subtasks:

\textit{Subtask 1 -- Source Detection:} 
Given a triplet (user query, reasoning trajectory, final answer), the participant systems should identify the source of the reasoning trajectory and final answer---whether they are generated by an AI system or written by a human.
The queries mainly involve math, coding, and real-life financial reasoning tasks.
This subtask supports a deeper comparison of reasoning style and cognitive structure between humans and models, helping to inform model-alignment and reasoning-training strategies.

\textit{Subtask 2 -- Safety Detection:} 
Given a triplet (user query, reasoning trajectory, final answer), where the query is from three categories --- (a)~risky queries requesting harmful content, (b)~jailbreak attacks with risks obscured by various strategies, and (c)~benign queries with risky tokens --- the participants must identify \Ni whether the reasoning trajectory is safe vs. unsafe, and \Nii whether the final answer is safe vs. unsafe.
This subtask targets step-level and outcome-level safety, to reveal unsafe reasoning hidden behind apparently correct conclusions.

For Subtask 1, in addition to existing datasets used to train reasoning for math and coding,\footnote{\url{https://huggingface.co/datasets/nivektk/math-augmented-dataset}} we plan to curate additional human-written reasoning trajectories with final answers from websites like \texttt{\small chegg},\footnote{\url{https://www.chegg.com/}} and collect AI-generated reasoning outputs using both cutting-edge open-source and commercial models such as Claude 3.7, DeepSeek-R1, Gemini 2.5 Pro, and GPT-5-Thinking~\cite{Claude3S, deepseekr1, gemini25,gpt5}, and other long-context or reflective LLMs.

For Subtask 2, several recent datasets provide stepwise and outcome-level safety annotations, such as STAIR~\cite{zhang2025stair}, Reasoning-to-Defend~\cite{zhu2025r2d}, and SaRO~\cite{mou2025saro}. We will leverage these corpora for training and validation, while 
the test set will be augmented with reasoning traces from the latest models featuring self-correction, reflection loops, and tool-use reasoning, to assess robustness under complex reasoning behaviors.

The evaluation of the Reasoning Trajectory Detection task will not be conducted inside the TIRA sandbox. Instead, we will use a separate evaluation platform where participants only need to submit their system outputs as files in the specified format.

Participants are free to develop and run their systems using any hardware or software resources of their choice, including local or cloud-based GPU acceleration, large open-weight models, or external APIs, as long as the final submission consists solely of the required output files. No code execution, model inference, or hardware constraints will be imposed at evaluation time.

This task bridges reasoning interpretability and safety assessment by evaluating not just what an LLM concludes, but how it reasons. It encourages systems that can detect unsafe or synthetic reasoning trajectories, quantify reasoning transparency, and explain their detection decisions, supporting the development of next-generation trustworthy reasoning models.

\section{Conclusion and Future Work}

We described the PAN 2026 shared tasks, which span robust AI detection, text watermarking, multi-author writing style analysis, generative plagiarism detection, and reasoning trajectory detection. Together, these tasks reflect the evolving landscape of AI-assisted text production and the need for rigorous evaluation frameworks that go beyond surface-level text characteristics.

Future work will focus on improving robustness, extending datasets across domains and languages, and refining the evaluation, to continue supporting research on trustworthy and reliable text analysis systems.

\begin{credits}

\subsubsection{\ackname} 
This work is partially supported by the European Commission under grant agreement GA\,101070014 (\url{https://openwebsearch.eu}) and by the Deutsche Forschungsgemeinschaft (DFG, German Research Foundation) -- 554559555.

\subsubsection{\discintname}
The authors have no competing interests to declare that are relevant to the content of this article.

\end{credits}

\begin{raggedright}
\small
 \bibliography{ecir26-pan-overview-lit}
\end{raggedright}

\end{document}